# Detection Of Concrete Cracks Using Dual-channel Deep Convolutional Network

Babloo Kumar
Department of Physics
National Institute of Technology
Durgapur

Sayantari Ghosh
Department of Physics
National Institute of Technology
Durgapur

*Abstract*— **Due to cyclic loading and fatigue stress cracks are generated, which affect the safety of any civil infrastructure. Nowadays machine vision is being used to assist us for appropriate maintenance, monitoring and inspection of concrete structures by partial replacement of human-conducted onsite inspections. The current study proposes a crack detection method based on deep convolutional neural network (CNN) for detection of concrete cracks without explicitly calculating the defect features. In the course of the study, a database of 3200 labelled images with concrete cracks has been created, where the contrast, lighting conditions, orientations and severity of the cracks were extremely variable. In this paper, starting from a deep CNN trained with these images of 256 × 256 pixel-resolution, we have gradually optimized the model by identifying the difficulties. Using an augmented dataset, which takes into account the variations and degradations compatible to drone videos, like, random zooming, rotation and intensity scaling and exhaustive ablation studies, we have designed a dual-channel deep CNN which shows high accuracy (~ 92.25%) as well as robustness in finding concrete cracks in realis-tic situations. The model has been tested on the basis of performance and analyzed with the help of feature maps, which establishes the importance of the dual-channel structure.**

## I. INTRODUCTION

Concrete structures, like towers, skyscrapers, dams and bridges become vulnerable to environmental exposures, and constantly weaken from use. One of the major surface defects that highlight urgent maintenance issues, are cracks. Cracks are developed due to thermal expansion & contraction, as well as caused by human damage which results into ultimate failure of any structure. Depending on the severity of the stress faced, cracks propagate towards stress concentrated areas e.g., discontinuity in the material or surfaces, sharp edges and deep corners [1].

To counter this problem, several administrative departments (like, Ministry of Road Transport & Highways, Federal Highway Administration etc.) inspect these structures on a regular basis. In this method of onsite manual inspection, cracks are identified by a qualified observer; the observer analyzes the crack image and gives the severity score to a particular crack. Manual inspection depends directly on skill and expertise level of the observer, and thus, is subjected to manual errors. Moreover, due to non-uniformly distributed environmental effects, the variety of cracks developed in a large-scale civil infrastructure needs thorough inspection, giving rise to accessibility issues [2]. Wastage of human resources and time are among other associated constraints of manual inspection due to which a constant inclination towards automated inspection is occurring. Identification, classification and evaluation of cracks, which is known as the problem of structural health monitoring, are being investigated with number of vision-based methods for detecting damages.

In the method of automated inspection [3], generally several images of civil structures, even from inaccessible angles, are capture by drone camera. These images are fed into robust algorithms based on fundamentals of image processing and computer vision to help analyzing the cracks automatically. Many hardware-based techniques rely more on the capture quality; though these techniques are good while dealing with large structures, these methods usually depend on dense instrumentations (e.g., numerous sensors, distributed sources, data integration) and get associated with problems like sensory system breakdown and noisy signals [4-6]. Image processing based methods are generally used in these tasks. Particularly in the area of concrete and pavement crack detection, Oliveria et al. [7] have detected the cracks based on entropy and image dynamic thresholding. Geodesic shadow removal [8], mathematical morphology through curvature evaluation [9], extensive preprocessing [10] and edge detection based algorithms [11] have also been used with success.

Through the last decade, machine vision algorithms, coupled with deep learning are showing immense prospect to solve classification and identification problems in several different fields [12-15]. Surveillance-based monitoring is becoming an area that is being constantly benefitted from deep-learning based methodologies. Convolutional Neural Networks (CNN) have been successfully employed for detecting fire outbreaks and fire disaster management [16, 17]. Surveillance of pavements and roads for pedestrian detection have been successfully executed by deep networks [18, 19]. Using deep CNN based computational methods, anomalous activities like theft, violence, stampede, overcrowding, intrusion etc. are being detected [20, 21]. These deep models are also being enhanced to deal with weather conditions and degraded surveillance footages.

The success of all these networks, indicate the necessity of creating deep-learning networks for structural health monitoring and crack detection using drones, where severe distortions can occur due to motion. In this paper, we propose a dual-channel CNN classifier based automated inspection method for crack detection which accomplishes classification task by training on a completely new dataset. The major contributions of this paper are:

- Preparation of an image dataset which will contain images of cracked and non-cracked surfaces from different concrete structures. This dataset will be annotated for the purpose of training, validation, and benchmarking for concrete structural health monitoring.
- A fast and reliable surface analysis for automatic crack detection based on a robust dual-channel deep CNN model, which can overcome the challenge of crack detection from noisy images and minimize the possibilities of manual error.

## II. METHODOLOGY

In our paper, the method consists of four steps: Image capture, database preparation, model training and model optimization. For image capture and database preparation, the steps followed are described in detail in respective subsections. Next, working with a goal of robust crack detection, we proposed a model that has deep sub-network structures. To appreciate the necessity of the subnetwork structures, we first discuss the single-channel classification network architecture, and demonstrate its restrictions. Next, we describe the architecture of Dual Channel Convolutional Network (DuCCNet) and implement the subnetworks with two parallel channels which contains skip conection and batch normalization.

### A. Image Capture

For dataset preparation, we captured relevant images with the help of a camera. All these images of walls, pavements, concrete roads were taken at National Institute of Technology, Durgapur campus. Images were captured in different daylight conditions (e.g., sunny, clear, cloudy and overcast days) and different timings during the day. In our case we have used a 16 MP camera without any zoom from a working distance of 1-3 m. The image resolution was 4068 × 3456 px. Each mother image represents a physical area of approximately 100 cm × 100 cm. Sample images are demonstrated in Fig. 1. In this entire processing, the surface illumination, contrast, color information, sharpness etc. were kept intact.

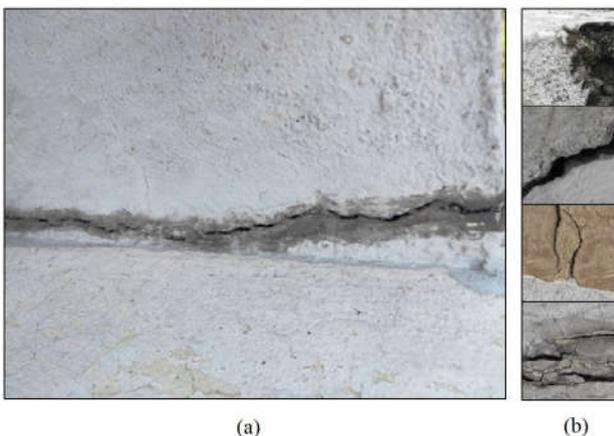

Fig 1:- Examples of (a) 4068 × 3456 px raw image captured with 16 MP camera containing both cracked and not-cracked region, and (b) some cropped-images of different crack types.

### B. Database Preparation

For preparing a standard database for the purpose of autonomous crack detection using deep networks, each of the mother images (4068 × 3456 px.) were divided into a set of 256 × 256-px cropped-images. The physical area of captured in these cropped-images is approximately 6 cm × 6 cm. The total number of cropped images are 3200 which contain both cracked and not-cracked images. All the cropped-images were labelled as cracked or non-cracked, and stored in the corresponding folder within the database directory [22]. The database became a versatile collection of labelled images which contains, smooth not-cracked surfaces, prominent cracks, thin cracks, shadows, crack-like rough surface textures, voids, edges, joints and color variations. These are shown in details in Fig. 2.

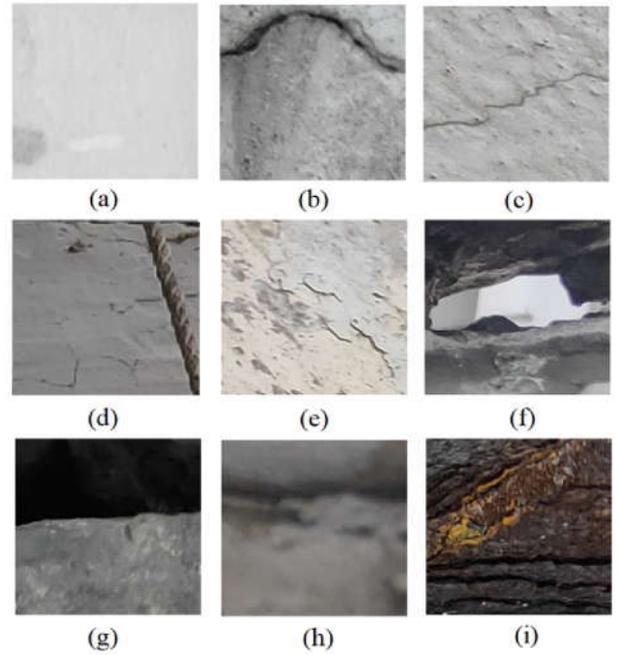

Fig 2:- Examples of 256 × 256 px cropped-images from contributed dataset which includes (a) Not-cracked surfaces, (b) prominent cracks, (c) thin cracks, (d) objects, (e) crack-like rough surface textures, (f) voids and (g) edges. The dataset also contains (h) degradations (like, blur) and (i) low contrast images.

### C. Single-Channel Network Architechture

Our primary network is a single-channel convolutional neural network (SCNN) which uses a series of convolutional blocks with intermediate max-pooling layers for feature extraction. For gathering local information and generating tensor of outputs, kernel convolution is an well-known tool. For a certain input image $I$ and our kernel $w$, subsequent 2D convolutional feature map values are calculated as,

$$G(m,n) = (I * w)[m,n] = \sum_i \sum_j w[i,j] I[m-i, n-j]$$

where * indicates convolution operation, and $m$ and $n$ indicate the indices of rows and columns of the result matrix respectively. To work with color images and for applying multiple filters in one layer, convolution over volume is necessary. Here the dimension of the output tensor changes obeying the equation below,

$$[N, N, N_c] * [K, K, N_c] = \left[\left\lfloor\frac{N+2P-K}{S}+1\right\rfloor, \left\lfloor\frac{N+2P-K}{S}+1\right\rfloor, N_k\right]$$

where, *N* and *K* are is the image and kernel size respectively, $N_c$ and $N_k$ denote number of channels in the image and number of filters respectively and *P* and *S* indicate used padding and stride respectively. The used SCNN srchitecture has seven convolutional blocks, where each block is made of three convolutionals layers. Each convolutional layer has 32 (size 3×3, stride 1, padding 'same') filters. To reduce the dimension of data and to impose the invariance learning, we affix a max-pooling operation of size 2 after each convolutional block.

Before the series of the convolutional blocks begin, the input image is introduced into the network through one single convolutional layer followed by a Batch Normalization (BN) transform [23]. During training of deep networks, constantly updating distribution of internal layer inputs affects the performance of the convolutional layers and training accuracy. BN handles this problem by transforming internal inputs. To enhance the stability of a neural network, this technique normalizes the output of a previous activation layer by subtracting the batch mean and dividing by the batch standard deviation. Then it adds two trainable parameters to each layer, γ and β, which are called standard deviation parameter and mean parameter respectively. The normalized output is multiplied by γ and added to β. Thus the input of the BN operation is the values of *x* in a mini-batch $\mathcal{B} = \{x_{1,2,...,n}\}$. If $m_B$ and $\sigma_B$ are mini-batch mean and standard deviation, then first the normalization is done as:

$$\hat{x}_i \leftarrow \frac{x_i - m_B}{\sqrt{\sigma_B^2 + \epsilon}}$$

Next, the trainable parameters are added as:

$$\hat{y}_i \leftarrow \gamma \hat{x}_i + \beta$$

Thus, minimization of the loss function is achieved by varying these two weights only, instead of changing all the weights, which might result into loss of the stability of the entire network. Moreover, by adding some noise to layer activations, BN also reduces overfitting through some minute regularization effect [23].

In the *k*[th] convolutional layer, after the convolution with the filter ($W^k$) and being added with a bias ($B^k$), a nonlinear activation function $\vartheta$ is applied to each feature map coming out of *(k-1)*[th] convolutional layer:

$$G^{[k]} = \vartheta^{[k]}(W^{[k]} \cdot G^{[k-1]} + B^{[k]})$$

Nonlinear activation functions allow the network to learn structural complexity of the data. We have used rectified linear activation unit (ReLU) activation for each convolutional layer:

$$\vartheta_{ReLU}(x) = \max(x, 0)$$

ReLU is identified as having close similarity to biological neurons where we expect that any positive value will be returned unchanged whereas a negative value will be set to 0 [24].

Once the features are extracted using the seven convolutional blocks, we append the output of the final convolutional block to a dense layer (ReLU activated) of size 32, after converting it to a column vector. We finally add a *sigmoid* activated dense layer of size 1 as the final layer of the network where the activation $\vartheta_{sig}(x)$ is defined as:

$$\vartheta_{sig}(x) = \frac{1}{1+e^{-x}}$$

Notably, all the activation functions are applied element-wise. As the final layer has to take decision about the classification based on probabilities, sigmoid function, being differentiable and existing between fixed range, works perfectly for this task. Here, we must mention that dense layers are basically fully-connected layers which are essential for classification tasks. However, full connections sometime cause over-fitting as they normally contain a considerable number of learnable parameters. We solve this problem by introducing dropout of 0.5 in between the dense layers. Thus, in each training iteration, instead of learning all the weights, the network continues learning a fraction of the weights. This SCNN model has total 159,201 trainable parameters.

TABLE I.   SYMBOLS AND COLORS USED IN NETWORK ARCHITECHTURE

| Symbols | Implications |
|---|---|
| 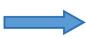 | Transmission |
| 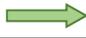 | Max pooling |
| 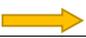 | Flatten |
| 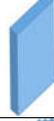 | Single Convolutional Layer |
| 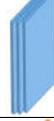 | Convolutional Block with three convolutional layers |
| 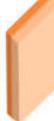 | Batch Normalization |
| 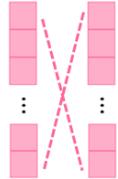 | Fully Connected Layers |

### III. RESULTS & ANALYSES

In this section we will discuss the performance of the proposed SCNN model and indicate its difficulties. Next, we will optimize the single-channel model with dual-channel convolutional network (DuCCNet) to increase stability and deal with vanishing gradient for detecting concrete cracks. We will demonstrate the performance of the model using standard parameters and feature maps. This section is comprised of the experimental procedure, model training, model optimization, model performance and ablation studies.

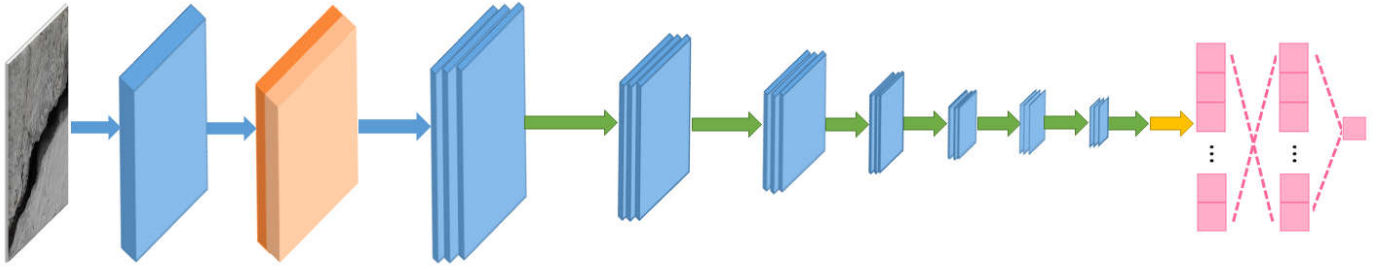

Fig 3:- The basic architecture of single-channel deep CNN (SCNN), as proposed in II-C. The significance of symbols and colors used is shown in Table I.

## A. Experimental procedure

For our entire experimental procedure, TensorFlow [25] and Keras [26] are used as platform. The experimental environment is the Windows 10 operating system running on a computer with an Intel(r) core i7-8550u CPU @ 1.80 GHz. The performance of the model has been estimated by validation accuracy (VA), which is defined as:

$$VA = \frac{N_C^D + N_{NC}^D}{N_C^T + N_{NC}^T} \times 100\%$$

where, $N_C^T$ and $N_{NC}^T$ are total number of with-crack and without-crack images, belonging to the test set, completely non-overlapping with training set. $N_C^D$ and $N_{NC}^D$ are number of images correctly detected by the network.

As this is a binary classification task, we use binary cross-entropy as the loss function. The loss function is given by:

$$H_{p,q} = -\frac{1}{N}\sum_{j=1}^{N}[x_j.\log p(x_j) + (1-x_j).\log(1-p(x_j))]$$

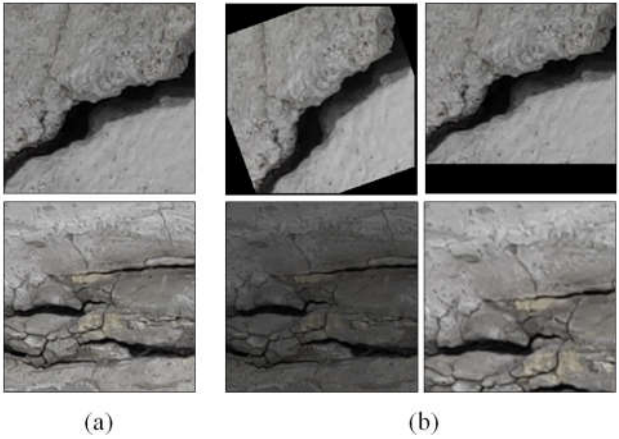

Fig 4:- Examples of data augmentation on $256 \times 256$ px cropped-images: Column (a) shows original iages from the training set while column (b) shows images rotated by $20^0$, shifted vertically, with scaled intensity and ramdomly zoomed.

where $x$ is the respective labels assigned, and $p(x)$ is the predicted probability of an image belonging to Class 1 (say, cracked) for all $N$ points. The minimization of loss function has been also recorded for testing the model performance.

## B. Model Training

To begin our study, we take the colour images of given concrete slabs from our dataset and resize it to 64×64 dimensional colour image; next, normalization has been performed. The Fig. 3 illustrates our initial model as described in Section II-C (Table I should be referred for interpretation of the symbols used in the figure). The model is trained using 6400 images with cross-validation split of 0.1 using ADAM optimiser. ADAM is a momentum-based optimiser that uses the squared gradients to scale the learning rate. It also takes advantage of momentum by using moving average of the gradient instead of gradient itself. We used this model with learning rate 0.0005, and default momentum parameter values. This model achieved 90.5% validation accuracy for available input images.

Though we attain a significantly high VA using the proposed network, investigations are performed to study the robustness of the proposed architecture. Dedicated efforts are going on to make structural health monitoring through automated drones to overcome the difficulties associated with manual accessibility and wastage of time. Drone images are often associated with problems like blurring, intensity variation, rotation, shift etc. Thus, it is important to make the model architecture robust against to different image degradations that may appear due to drone vibrations, motion blurring, lack of expertise, poor lighting condition, and lack of camera resolution.

Thus, to enrich our data and to strengthen our network, we performed rigorous data augmentations. We applied random rotations (up to $25^0$), width and height shift (up to 10% of the image size), random zooming (up to 20%), intensity scaling (up to 20% of the original intensity) and horizontal and vertical flip on the images of the dataset. On this augmented dataset, this SCNN gave VA of 82.25% using 0.1 cross-validation. This kind of fall in performance under image degradation and data augmentation has been reported before for deep networks [27, 28].

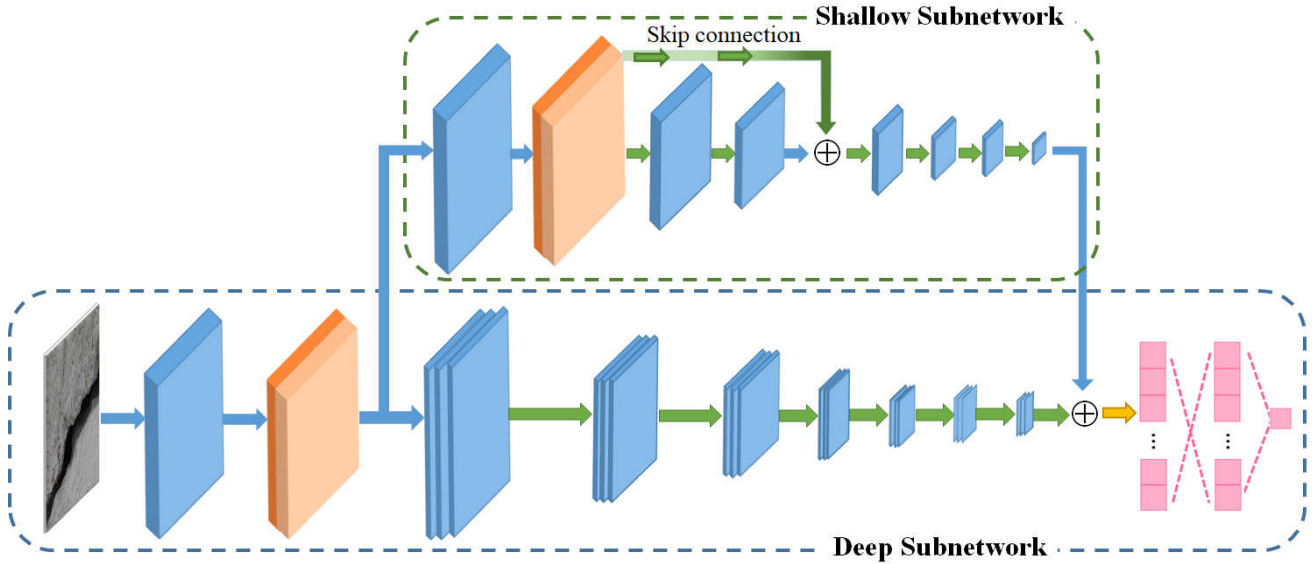

Fig 5:- The basic architecture of dual-channel convolutional network, namely DuCCNet, including twenty nine convolutional layers, sixteen max-pooling operations, two batch normalizations and two fully connected layers. The meaning of symbols can be found in Table I.

## C. Model Optimization

Detailed and exhaustive experiments with proposed SCNN established the fact that the network performs extremely well for extracting the detail of crack information. But, it suffers from vanishing gradient problem. The SCNN in total is constituted of seven convolutional blocks having 21 convolutional layers. As the number of layers increases in the neural networks, the gradients of the loss function approaches to zero, making the network hard to train. Being a very deep network with 21 convolutional layers, this problem of exponential fall of gradient increases through back-propagation. To overcome this problem, a second channel is added to the SCNN, giving rise to a dual-channel convolutional network with subnet structure. This second channel has two major properties, which can contribute to deal with the vanishing gradient problem: comparatively shallow network structure and skip connection.

**Shallow Network structure:** In strict nomenclature, a shallow neural network will have just one hidden layer. Here, however, we refer to this added second channel as shallow, to indicate the contrasting number of total convolutional layers, compared to the SCNN. The portion of the SCNN which lies in parallel with the second channel has 21 hidden layers, while that of this second channel has only 7 layers.

**Skip connections (SC):** With the increase in depth, problems related to convergence arises in neural networks. We follow the approach of Deep Residual Learning [29] where connections from the output of one layer gets added with the input of a much earlier layer, skipping some of the intermediate convolutional layers in process. This process gives rise to a residual function which is easier to optimize than the original mapping.

Moreover, after the first convolutional layer of the second channel, a BN transformation has been done prior to the max-pooling, to increase stability.

**Dual-Channel Convolutional Network (DuCCNet):** The detailed structure of the modified network that introduces the benefits of shallow structure, SC and BN in the architecture can be seen in Fig 5. The input image is to be pushed into the network using the first convolutional layer followed by a BN operation. After BN, the network gets bifurcated into two subnetworks. Here, we must note that the channel below is identical with the SCNN we were using before (as seen in Fig. 3). The entire deep structure with all the convolutional layers, max-pooling connections and dense layers are exactly same as before. This is referred as Deep-subnetwork (Deep-SN) of the network.

The top-channel, on the other hand, is the comparatively shallow channel with seven convolutional layers. This second channel is referred as shallow-subnetwork (Shallow-SN) of the complete network. Once the bifurcation occurs, the input is first convoluted and batch normalised once again. Each convolutional layer has 32 filters, each of size 3×3, stride1, and padding 'same'. Here also we have used ReLU activation for each convolutional layer. The number of max-pooling remains the same (six max-pooling of size 2) with Deep-SN to maintain the correlation between sizes. Two convolutional layers (second and third) of Shallow-SN are skipped using a skip connection to make it further shallow, if needed. The skip-connection only contains two max-pooling connections of size 2. After the seventh convolutional layer, the output of the Shallow-SN is added to output of the seventh convolutional block of Deep-SN. This added output is flattened and passed through the ReLU activated dense layers with a dropout of 0.5 as before for the final classification. This complete model is referred as Dual-Channel Convolutional Network (DuCCNet). The model has 233,441 trainable parameters. Thus, by adding the Shallow-SN, the capacity of the entire network has also been enhanced.

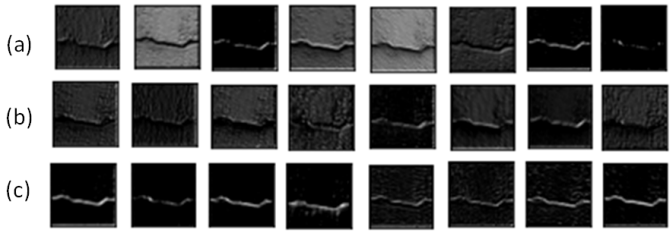

Fig 6:- Convolutional feature maps (CFM) from different parts of the network. Selected CFM (Eight each) can be seen for (a) before the bifurcation of the two channels, (b) after the bifurcation and after the first convolutional layer of Deep-SN, and (c) after the bifurcation and after the first convolutional layer of Shallow-SN. Each feature map (a), (b), and (c) exhibits learned features detected across the image.

### D. Model performance

With cross-validation split of 0.1, the model is trained using 6400 images using ADAM optimiser with learning rate 0.0005 and default momentum parameter values. The model gives a VA of 92.25% with the same augmented data-set. Average time taken for each epoch is 159.031 seconds. To examine model performance, first we visualize the feature maps from both the sub-networks to ensure their complementarity and necessity. We perform a detailed study of extracted feature maps prior to each pooling layer from different levels of the sub-net-structures. Some selected feature maps for before and after the bifurcation of the channels are shown in Fig 6. We note that while the Deep-SN detects minute features from the input image (Fig. 6 (b)), the Shallow-SN learns the prominent cracks (Fig. 6 (c)). By adding these features together DuCCNet enables us to detect cracks from all categories with increased accuracy.

Next, we show the curves of training accuracy and validation accuracy to illustrate the whole training process in Fig. 7(a) which shows that DuCCNet has a very stable training process, as the variation of VA and training accuracy was very consistent. Fig 7(b) shows the variation of training loss and validation loss, which ensures absence of overfitting, as both curves are simultaneously decreasing.

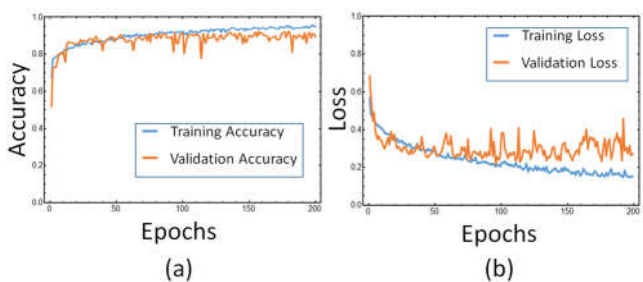

Fig 7:- Performance of the model in terms of convergence, stability and overfitting:. (a) Training and Validation accuracy (VA) as well as (b) Training and Validation losses for DuCCNet are shown.

### E. Ablation studies

We also report that, while optimizing the model architecture, we have performed several ablation studies to finalize the model. To understand the results, we report the performances of five different models in Table II. In the table, 'Model 1' and 'Model 2' denote SCNN without and with the final convolutional block. In Model 1, the $6^{th}$ convolutional block is flattened and connected with the dense layer. In 'Model 3' we added the skip-connected second channel with Model 1, and in 'Model 4' we added the second channel without the skip connection to Model 2. Finally, the proposed DuCCNet is designed by adding skip connected second channel to Model 2. Among all the variations that we explored in our ablation study, proposed DuCCNet outperforms all other variants. This further justifies the necessity of the Shallow-SN and the skip-connection.

TABLE II. ABLATION STUDY OF THE PROPOSED MODEL

|  | Model 1 | Model 2 | Model 3 | Model 4 | DuCCNet |
|---|---|---|---|---|---|
| **Channel 1** | ✓ | ✓ | ✓ | ✓ | ✓ |
| **Channel 2** | ✗ | ✗ | ✓ | ✓ | ✓ |
| **Skip Connection** | ✗ | ✗ | ✓ | ✗ | ✓ |
| **Conv-block7** | ✗ | ✓ | ✗ | ✓ | ✓ |
| **Accuracy** | 79.75 | 82.50 | 85.75 | 89.00 | ***92.25*** |

### IV. CONCLUSION AND DISCUSSIONS

Automated inspection gives highly accurate result as compared to the slower subjective manual inspection procedures. In the present study, a vision-based method for concrete crack detection using a dual-channel CNN is proposed. The backbone of this network is based on capturing basic as well as minute details of the image using a combination of shallow and deep neural networks. For overcoming overfitting and vanishing gradients, repetitive batch normalizations and skip connection have been introduced.

With this network, extensive experiments have been carried out to detect concrete cracks. The dataset prepared and annotated for the purpose of the experiments are made freely available to the research community [22]. The network gives an accuracy of 92.25% with an augmented dataset which comprises of rotated, shifted, randomly zoomed and intensity scaled images of the original images, which establishes excellent robustness of the network. We also monitor the training process by tracking evolution of accuracies and losses, which shows fast convergence, stability and absence of overfitting. Extracted feature maps and ablation studies support the optimal structure of the network.

This network can find wide application in the area of structural health monitoring due to its robustness and stability. This architecture, namely DuCCNet, and close variants of it look extremely promising for drone-based civil structure surveillance, as it has been optimized to counter the presence of random rotations, zooming and intensity scaling, which occur very frequently in drone-camera recorded video. In a future study, this research will be extended to detect the type, number, width and length of cracks on the structures and classifying the degradation level, based on the carrying capacity of the concrete structure. Moreover, further studies will be carried out to find improved classifiers based on this architecture, and its hardware implementation.